# Object Tracking based on Quantum Particle Swarm Optimization

Rajesh Misra
Lecturer at S.A.Jaipuria College
rajeshmisra.85@gmail.com

Kumar S. Ray
Professor at Indian Statistical Institute
ksray@isical.ac.in

Abstract –In Computer Vision domain, moving Object Tracking considered as one of the toughest problem. As there so many factors associated like illumination of light, noise, occlusion, sudden start and stop of moving object, shading which makes tracking even harder problem not only for dynamic background but also for static background. In this paper we present a new object tracking algorithm based on Dominant points on tracked object using Quantum particle swarm optimization (QPSO) which is a new different version of PSO based on Quantum theory. The novelty in our approach is that it can be successfully applicable in variable background as well as static background and application of quantum PSO makes the algorithm runs lot faster where other basic PSO algorithm failed to do so due to heavy computation. In our approach firstly dominants points of tracked objects detected, then a group of particles form a swarm are initialized randomly over the image search space and then start searching the curvature connected between two consecutive dominant points until they satisfy fitness criteria. Obviously it is a Multi-Swarm approach as there are multiple dominant points, as they moves, the curvature moves and the curvature movement is tracked by the swarm throughout the video and eventually when the swarm reaches optimal solution , a bounding box drawn based on particles final position. Experimental results demonstrate this proposed QPSO based method work efficiently and effectively in visual object tracking in both dynamic and static environments and run time shows that it runs closely 90% faster than basic PSO.in our approach we also apply parallelism using MatLab 'Parfor' command to show how very less number of iteration and swarm size will enable us to successfully track object.

## I. Introduction

Object Tracking simply employs the idea of following an object as long as its movement can be captured by a camera in various environment like either camera is moving with the object(Variable Background) or Camera is static (Static Background).moving object detection and tracking pose a challenge in real world scenario like surveillance system, traffic monitoring, vehicle navigation etc. In many scenarios where background changes dynamically due to motion of camera, abrupt changes in speed of the tracked object, illuminating noise, occlusion creates tracking more challenging. Therefore tracking algorithms under such situation should be robust, flexible and adaptive and capable of real time execution.

Moving object tracking is a more than a two decade problem, so many methods has been proposed with a certain degree of effectiveness. Considering bio inspired swarm based method as effective tools for object tracking draws extensive attention in past decades. Among other bio inspired method like Genetic Algorithm (GA), Particle Swarm Optimization (PSO) emerges real fast because of its efficient, robust and quick convergence. Some of the earlier work has been done on tracking using PSO successfully.

Particle Swarm Optimization is applied by Zheng and Meng[14] on high dimensional feature space for searching optimal matching in Haar-Like features detected by a pre-defined classifier set. Xiaoqin Zhanget all in [15] calculates temporal continuity between two frames and use that information for swarm particle to fly and track that information. Vijay John, Emanuele Trucco, Spela Ivekovic [13] construct Human Body Model as a collection of truncated cones, numbering those cones and PSO cost function will check how well a pose matches with data taken from multiple camera. Multiple people tracking has been done by Chen Ching-Han and Yan Miao-Chun [11] where people body constructed by feature vector and histogram and PSO used that histogram information for its tracking purposes.Fakheredine Keyrouz shows in [6] use of multiple swarm for multiple parts of object tracking and those swarm will share information with each other to make tracking of object as a whole. Multiple object tracking also be done by Chen-Chien Hsu, Guo-Tang Dai [9] using PSO, there method is constructing a feature model using grey-level histogram and apply PSO particles to track the difference between grey level histogram information of consecutive frames in a video sequence. Bogdan Kwolek [5] represent an approach where object is represented by image template, after constructing a covariance matrix based on that, by using similarity measure PSO will keep track of the difference appeared during movement of the object with target template.

Above we discussed all those considerable contribution towards object tracking using PSO in various ways, but there are significantly less work has been done in object tracking using Quantum Particle Swarm Optimization (QPSO), Though this QPSO concept is not very old[17] developed by Sun, Jun, Bin Feng, and Wenbo Xu but still a decade gone, while we found Chen, Jinyin, Yi Zhen, and Dongyong Yang work [4] on object tracking but other than that not much work to mention.

In this paper we approach a new method for object tracking under different background scenario. The uniqueness of our algorithm lies on three factors,1) use of Dominant points 2) a singular algorithm applicable for static and variable background. 3) use of Quantum Particle Swarm Optimization method. None of the earlier work discussed above use dominant point as tracking tools except Prasad, Dilip K., and Michael S. Brown[4] uses dominant point for boundary contour and propose a time propagation method of those dominant points so that it can effectively detect deformation of object and track. Our method completely different than this approach, we calculate chain code, chose dominant points, determine the curvature between two dominant points, randomly define swarms and let the swarms track that curvatures from frame to frame.The use of quantum PSO helps us to reduce swarm size and number of iteration hugely, as a result our algorithm takes less time compare to other PSO approach.

This paper organized as follows. Section II discusses basic QPSO algorithm, Dominant Point and Optical Flow method. Section III cover detail approach of our proposed algorithm. Section IV cover how we implement QPSO using parallelism and Section V gives experimental results and in Section VI we will discuss overall performance and comparison between PSO, QPSO and QPSO parallel algorithms and finally we draw conclusion.

## II. Quantum Particle Swarm Optimization

### a. Particle Swarm Optimization

In 1995 James Kennedy and Russell Eberhart proposed an evolutionary algorithm that create a ripple in Bio-inspired algorithmic approach called Particle Swarm Optimization (PSO). In a simple term it is a method of optimization for continuous non-linear function. As this method influenced by swarming theory form biological world like fish schooling, bird swarming etc.

PSO effectively applied to the problems in which each solution of that problem can be consider as a *set of points* in a solution space. *Particle* is the term associated to those set of points. Analogically suppose there is a food source and a swarm of birds tries to reach that food source. Every bird will try by its own to reach there and whomever is reached or nearly reached to that food source will share that information with other birds who are close neighbor, as a ripple in water that information will be flown among entire swarm of birds and every birds will synchronously update their velocity and position if they got better position in terms of nearest position to the food source .As a result after certain period of time entire swarm will eventually gather to the food source. Similarly every solution considered as particle will compute there value based on some cost function, until they satisfy certain criteria known as *stopping condition*, they will keep updating their velocity and position, if their neighbor got better solution.

Position and Velocity are two associated terms in Particle Swarm Optimization. Position of every particle is calculated by particle's own velocity. Let $X_i(t)$ denote position of particle i in the search space at time t. position Updation formula is as follows –

$$X_i(t+1) = X_i(t) + V_i(t+1) \qquad (1)$$

Where

$V_i(t+1)$ is the velocity of particle i at time (t+1), which will be computed based on this following formula –

$$V_i(t) = V_i(t-1) + C_1 * R_1 (P_{LB}(t) - X_i(t-1)) + C_2 * R_2 (P_{GB}(t) - X_i(t-1)) \qquad (2)$$

Where

$C_1, C_2$ = Constants determine the relative influence on social and cognitive components, also known as *learning rate*, often set to same value to give each component equal weights.

$R_1, R_2$ = random values associated with learning rate components to give more robustness.

$P_{LB}$ = Particle Local Best position – it is the historically best position of the i[th] particle achieved so far

$P_{GB}$ = Particle Global Best position – it is the historically best position of the entire swarm, basically position of a particle which achieve closest solution.

Equation (2) is Kennedy and Eberhart's original idea, after that lot of different research has been going on, based on that one of the remarkable idea comes up by Shi and Eberhart [15] of addition of a new factor called "inertia weight" or "w ". After addition of inertia weight the eq. (2) becomes as follows –

$$V_i(t) = w * V_i(t-1) + C_1 * R_1 (P_{LB}(t) - X_i(t-1)) + C_2 * R_2 (P_{GB}(t) - X_i(t-1)) \quad (3)$$

This inertia weight helps to balance local and global search abilities, small weight means local search and larger weight means global search.

### b. Quantum Behaved Particle Swarm Optimization :

The main disadvantageous factor of basic PSO algorithm is Convergence. It is not guaranteed that it will converge or not after certain number of iteration. There are many different approaches has been taken by different researcher like Hybrid PSO, Variable PSO, GA-based-PSO to tackle this issue. Quantum PSO is another such approach with a different angle which not only guarantee convergence but also assure its speediness over basic PSO.

In the quantum world the state of a particle is determined by its wave function $\Psi_{(x,t)}$ instead of position and velocity because according to Heisenberg's "*Uncertainty Principal*" we cannot determine position and velocity of a particle at the same time. We will calculate probability of particle position in x using probability density function $|\Psi_{(x,t)}|^2$ . In Quantum time space framework the particle will move according to the following iteration introduced by Jun Sun *et al.*

$$X_{(t+1)} = P_i - \beta * (mBest - X_t) * \ln(1/u) \text{ if } K \geq 0.5 \quad (4)$$

$$X_{(t+1)} = P_i + \beta * (mBest - X_t) * \ln(1/u) \text{ if } K < 0.5 \quad (5)$$

Where,

$$P_i = \varphi * pBest_i + (1 - \varphi) * gBest \quad (6)$$

$$mBest = 1/N \sum_{i=1}^{N} pBest_i \quad (7)$$

in the formula (7) mBest is the current centre of all individual optimal location which is the mean best position of all the best position of the population. K , u , φ are random number distributed uniformly over the range [0,1]. ß is the only different parameter used in QPSO known as Contraction - Expansion coefficient. It will be used to control convergence speed by tuning its value. N is the population size. pBest and gBest aare same as basic PSO's $P_{LB}$ and $P_{GB}$.

We can observe the difference between PSO and QPSO as QPSO introduce stochastic distribution of particle position. This exponential distribution makes the search space a real space covering whole solution space which increase the chances of optimality, besides introduction of mBest enhances the convergence possibility.

### QPSO algorithm

Pseudo code of the QPSO algorithm as follows –

### Procedure QPSO
For each particle
   initialize particle $X_i$ by randomizing them;
Evaluate $X_i$ ;
$pBest_i = X_i$ ;

END

Do

Compute gBest = Min or Max ( f(pBest));
Compute mBest by equation (7)

   For each particle

      Calculate $P_i$ using equation (6)
      Update Position $X_i$ using equation (4 and 5)
      If $f(X_i) < f(pBest_i)$
      $pBest_i = X_i$ ;
      ENDIF
EndFOR

Until While maximum iterations or minimum error criteria is not attained

### c. Dominant Points

F. Attneave observation [25] on information about shape and overall structure of any curve lies on those points having high curvature. Since then dominants points are considered as one of the significant candidate of object boundary detection because they hold important feature information about object contours. There are number of approaches has been offered by various researchers on efficient, successful detection of dominant points after Attneave's careful observation.

According to Wu, Wen-Yen[18], A digital curve C consist of n consecutive points and can be written as follows –

C = { $p_i(x_i, y_i)$ | i = 1,2,3...n}

$p_i$ is the $i^{th}$ point having coordinate values $(x_i, y_i)$.

As Ray and Ray proposed k-Cosine value [22] as follows –

$$(a_{ik}.b_{ik})$$

$$Cos_{ik} = \frac{\text{-----------------}}{|a_{ik}| \cdot |b_{ik}|} \qquad (8)$$

Where vector $a_{ik} = (x_{i-k} - x_i, y_{i-k} - y_i)$, $b_{ik} = (x_{i+k} - x_i, y_{i+k} - y_i)$, $\cdot$ is the inner product operator. And $|*|$ is the vector length. After calculating k-cosine values by above formula (3) of each boundary points the dominant points will be selected as those points having maximum k-cosine values. This k-cosine values is used to determine the length of the support region. Maximum length of the support value will give higher chance of the particle for selection of dominant points. We use Ray and Ray [22] method and Wu, Wen-Yen [18] method for dominant point selection which will be discussed in next section in detail.

### d. Optical Flow method

According to Horn, Berthold KP [24], optical flow is "distribution of apparent velocities of movement of brightness patterns in an image", in other terms it is the change of the light in an image due to motion of camera sensors. A video can be thought of as a sequence of image frames, which can be constructed from spatial and temporal sampling of incoming light. Then optical flow can captures the light changes in these images using vector fields.an accurate, pixel wise estimation of optical flow gives correct position of pixels in consecutive image sequences.

There are different methods for determining optical flow like Horn–Schunck method, Lucas -Kanade method. We use Lucas -Kanade optical flow method in this paper which is discussed next.

### e. Lucas -Kanade method optical flow method (KLT)

The basic optical flow problem is How to estimate the motion form one image frame to next one? Lucas- Kanade method assumes the motion of the pixels nearly constant in local neighborhood of the pixel under considerations. In mathematical form it will looks like as follows –

$$I_x(q_1) V_x + I_y(q_1) V_y = -I_t(q_1)$$
$$I_x(q_2) V_x + I_y(q_2) V_y = -I_t(q_2)$$
$$\vdots$$
$$I_x(q_n) V_x + I_y(q_n) V_y = -I_t(q_n)$$

where $q_1 \ldots q_n$ are pixels inside the window, $I_x(q_i), I_y(q_i), I_t(q_i)$ partial derivatives of the image I with respect to position x, y and time t, evaluated at the point $q_i$ at the current time.

This formula can be written as matrix form $Av = b$

$$A = \begin{bmatrix} I_x(q_1) & I_y(q_1) \\ I_x(q_2) & I_y(q_2) \\ . & . \\ I_x(q_n) & I_y(q_n) \end{bmatrix} \quad v = \begin{bmatrix} V_x \\ V_y \end{bmatrix} \quad b = \begin{bmatrix} -I_t(q_1) \\ -I_t(q_2) \\ . \\ -I_t(q_n) \end{bmatrix}.$$

This mathematical system has more equation than unknown, Lucas-Kanade applies Least Square principle to solve this equation. As follows –

$$A^T A v = A^T b.$$

The matrix $A^T A$ is known as structure tensor of the image at a certain point.

### III. QPSO Based Tracking Approach

#### A. General Idea

We begin our tracking process by first calculating dominant points of object. As dominant points hold the highest curvature values of the object so selection of dominant points will definitely be a critical task. in our approach we took a random number which will guide how many dominant points will be selected from the object body and for calculating dominant points we follow [22][18]] approach.

After we find required number of dominant points we create a swarm of random particle by using (1) and (2). Those swarm will find its closest curvature that will connected by two consecutive dominant points, as per tracking purpose those swarm will track their curvature as long as object moves. From frame to frame those dominant points are tracked by Lucas –Kanade ( KLT) optical flow algorithm. So every time a new frame appears old dominant points change their position and new position will be calculated by KLT method and this new-position curvature will again tracked by QPSO algorithm.

As we can see there are multiple curvature of a single object body, and each one tracked by an individual swarm so there will be multiple swarm for object tracking. In each frame once the QPSO successfully finds all curvature connected by dominant points, we will create a bounding box around the moving object. For creation of that bounding box we didn't follow any predefined bonding box algorithm rather we formulate our own algorithm which will suite our QPSO based method most.

#### B. Dominant Point selection

On Dominant point detection we first perform contour tracking of the target object body to find the Chain Code. For that purpose we use Freeman Chain Code. Freeman Chain code gives us list of pixels around object body. Among those pixels we eliminate linear points, as those points does not provide us any significant curvature information. For elimination of linear points we follow the following rule –

If $C_{i-1} = C_i$ then point $P_i$ is linear point. (9)

Where $C_{i-1}$ is previous chain code value and $C_i$ is current one on point $P_i$ .

After excluding those linear points rest of the points are called breakpoints, which are candidates for dominant points. We have to consider region of supports of only those breakpoints. Now we need to calculate length of support of each breakpoints. Rather considering all breakpoints at once we group them as a group of 10 for Variable background and group of 5 for static background. The number of breakpoints in a group will be decided based on which background we perform our tracking, normally on variable background object shape change fast for that we need our curvature of the object body smaller such that more breakpoints are close to each other that's why we chose high number of breakpoints, comparatively in static background as the object is more stable we can use much longer curvature so less number of breakpoints will suffice for dominant point calculation.

For each group of breakpoints we calculate k-Cosine values of each of them and apply following rule –

Start with k =1 form group. Increase value of k by 1 until we reach all breakpoints on this group. $k_i = k$ if $\cos_{ik} = \max \{ \cos_{ij} \mid j= K_{min} \ldots\ldots\ldots K_{max}\}$ for j = 1 , 2, …n  (10)

We chose dominant point as those points which are max k –Cosine values.

$D_i = \max \{ \cos(p_i)\}$  (11)

Whole procedure for calculating dominant points are as follows –

Step 1: Use Freeman Chain Code for performing couture tracking, get those pixels store them in a file.

Step 2: Eliminate linear points by following rule (9) form those stored pixels. Save them in a file called breakpoints

Step 3: Perform K –Cosine for each of the breakpoints by following rule (10).

Step 4: Select those points as dominant points which has max k-cosine values called Dominant point set.

C. Dominant point tracking by KLT method

The main advantage of dominant point is that they hold most curvature information which hardly changes if the environment changes, rather calculating dominant point every time on a new image frame it is always better if we can track those dominant points form frame to frame. For tracking those dominant point we apply optical Flow the method.

As earlier we discussed Lucas-Kanade Optical method which quite a useful approach for tracking any point in dynamic background. We run KLT method for each Dominant points and store their position $(X_i,Y_i)$ in separate table. So every time a new image frame comes, KLT algorithm gives us the new probable position.

D. Curvature tracking by QPSO.

What we assume as a curvature is that the path between two consecutive dominant points. As we know dominant points are boundary points so curvature connecting two dominant points gives us the object boundary curve. Now using QPSO we are going to track that curvature.

i) Setting QPSO parameters and Initialization

Because of dynamic nature setting QPSO parameters to right value is a crucial task.

- **Multiswarm** – our approach is based on multiple swarms and the number of swarm will be decided by how many dominant points we are considering. if we have D number of dominant points and each curvature will be represented by 2 dominant points then there will be –
  number of curvature (C) = D/2

  as each curvature is tracked by each swarm then there will be C swarms. It is worth considering that this is the maximum value of swarms we started with, as the procedure goes on it may be possible that some swarm may lost tracking because dominant points are not tracked by KLT method which is inherent problems in optiocal flow method. So latter no of swarm may reduce which will not affect much in our approach. as per our exoeriment we obseve that probably 10% of swarm may lost their path. So we can conclude that –

  10% of C <= no. of swarm <= C

- **Swarm size**- Here we have huge improvement over basic PSO. We initialize swarm size through some trial and error process and we conclude that for static background swarm size =7 and for variable background that will be 10, this are approximated values so if we set swarm size near about10 (for both static and dynamic environment) this algorithm works at its best. We can easily observe that we need very less number of particle in a swarm. It is obvious that more particles in a swarm means more computation and also less particle means divergence of tracked object, so we need to keep the swarm size optimum which is depend on different application. In our case it is depends on the length of the curvature, the long is the curvature the more particle we need to track that curvature successfully, but as far QPSO concerned we don't need high number of particle in a single swarm. Our experimental shows how very less number of particle do the job.

- **Position initialization** –As ofQPSO methodology we need to initialize the position of every particle of the swarm. This position of particle for each swarm will be inside the search space and randomly defined. In our case we first calculate the X and Y length of the image which is basically the dimension of the target image. Position of each particle is defined randomly on the [X,Y] range for each image frame.

- **Local best value of Particle i (pBest$_i$ )**– local best value of an individual particle in a swarm indicates its current best position it achieved in convergence with the target curvature. We initialize each particle Plbest value with its current position. Latter it will be modified according to the updating rule.

- **Global best value of Particle i (gBest )** – In a particular swarm, the particle which hold the best position such as closest to the curvature boundary considered as global particle and its position is gBest. each particle first compute the perpendicular distance from the curvature connected by dominant points, the particle which hold minimum distance considered as gBbest
  gBest = { min ( distance($X_{D1}$,$Y_{D1}$,$X_i$,$Y_i$,$X_{D2}$,$Y_{D2}$))}
  Where ($X_{D1}$,$Y_{D1}$) is the position of the 1st dominant point and ($X_{D2}$,$Y_{D2}$) position of the 2nd dominant point.

  - **mBest computation** – we will compute mBest according to the formula (7) once we completed our pBest and gBest calculation.

  - **φ ,ẞ, u , k values initialization** – earlier we define the meaning of this termson equation [4-6], all those variables values are randomly distributed in range [0-1].

  - **Position Updation** – particles position will be updated based on formula (4) and (5). If random value of k is greater than 0.5 then we will apply formula (4) else position will be calculated based on formula (5). Our computation reduced as there is no velocity concerned in QPSO.

II) Curvature Computation –

In this paper we construct the curvature based on dominant points. Let consider two dominant points are $D_1$ and $D_2$ calculated using formula (11). The curvature between this two points will be the curve joining this two points. There could be infinitely many curve that will pass through this two points, but in this paper we consider Euclidean Distances between this two points.
In Cartesian coordinate, $D_1$ ($X_1$,$Y_1$) and $D_2$($X_2$,$Y_2$) are the two points in Euclidian space, the distance between this two points will be calculated based on Pythagorean formula as follows –

$$\overline{D_1D_2} = \sqrt{(X_2 - X_1)^2 + (Y_2 - Y_1)^2} \qquad (12)$$

Using a random curve we can show as follows –

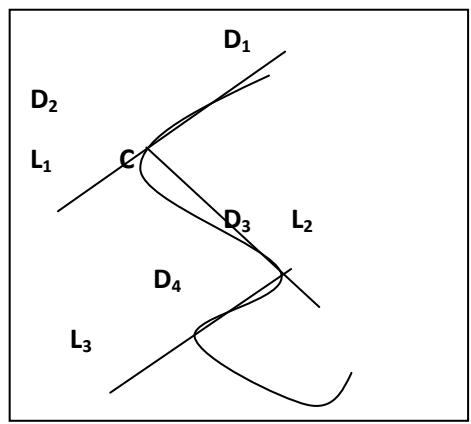

Figure – 1

In the above figure (1), we draw a curve C, which contains dominant points like $D_1, D_2, D_3, D_4$, line $L_1, L_2$ and $L_3$ passing through $D_1D_2$, $D_2D_3$ and $D_3D_4$ respectively. Though it is not exactly the curve connecting dominant points $D_1D_2$, $D_2D_3$ or $D_3D_4$ but as figure shows it serves the purpose of identifying approximate object boundary. As we are not detecting or tracking exact object body curvature, we are focusing only moving area of target so inexact curvature of the boundary not so serious threat for tracking. As it is always acceptable if we can find exact curvature which is another research area of couture tracking, not the main focus area in this paper.

III) Fitness Function

Every QPSO model based on some cost function, each particle of the swarm will compute that fitness function in each iteration to confirm whether they converge to the final solution or not. In this paper, our cost function is perpendicular distance of the particle i to the curvature under tracking

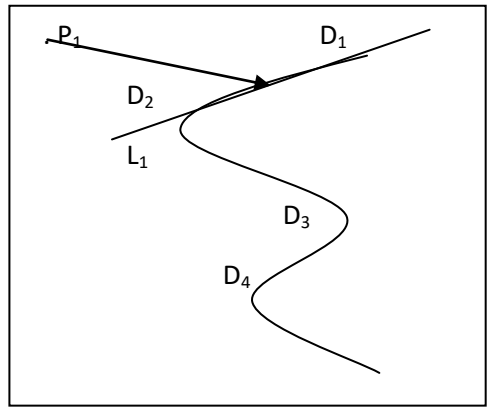

Figure – 2

Borrowing our earlier figure (1), we draw figure (2), here there is a particle $P_1$ and the curvature is line $L_1$. We are computing the perpendicular distance from the point $P_1$ to the line $L_1$.

As $L_1$ passes through two dominant points $D_1(X_1, Y_1)$ and $D_2(X_2, Y_2)$ then the distance from the point $P_1(X_0, Y_0)$ is –

$$\text{PerDist}(D_1, D_2, P_1) = \frac{|(Y_2 - Y_1) * X_0 - (X_2 - X_1) * Y_0 + X_2 * Y_1 - Y_2 * X_1|}{\sqrt{(Y_2 - Y_1)^2 + (X_2 - X_1)^2}} \quad (13)$$

The denominator is the length of $D_1$ and $D_2$. Numerator is the twice the area of triangle with its vertices at 3 points $D_1, D_2, P_1$.

Every particle will compute this perpendicular distance with curvature and if the distance is in the acceptable range iteration stops else this procedure continues. We summaries algorithm for cost computation as follows –

Procedure FitnessComputeQPSO (Particle set)
    for each particle $P_i$
        Compute PerDist $(D_1,D_2,P_i)$
        if PerDist $(D_1,D_2,P_i)$< acceptable range
        particle $P_i$ accepted
    end for
endFitnessComputeQPSO

    IV)    New pBest , gBest Updation and re-initialization

Until all the particle inside a swarm are successfully converged on the curvature they keep updating their position using formula (4) and (6). gBest and pBbest will be updated as –

New_gBest $(P_i)$ = {min (PerDist $(D_1,D_2,P_i)$) for all particle i }     (14)

New_pBest $(P_i)$ = $\begin{cases} \text{(PerDist } (D_1,D_2,P_i)) \text{ if (PerDist } (D_1,D_2,P_i))< \text{ previous (PerDist} \\ \quad\quad\quad\quad (D_1,D_2,P_i)) \\ \text{previous (PerDist } (D_1,D_2,P_i)) \end{cases}$     (15)

Reinitailization is sometime required as it is inherent nature of QPSO that sometime particles are too diverge that several updation may not bring them toward their goal, in our case it is also possible that some particle are too far away from curvature boundary and after a finite number of iteration they still unable to converge then we need to reinitialize those particle. Reinitializing particles over entire image space certainly feasible but not a practical idea, because it again may diverge, so we have a better possibility to converge if we assign the position randomly over the range of two particle's position which has the best position so far. mathematically we can write as –

Pos_x$(P_i)$ = rand(pos_x$(P_{i-1})$ , pos_x$(P_{i+1})$)     (16)
Pos_y$(P_i)$ = rand(pos_y$(P_{i-1})$ , pos_y$(P_{i+1})$)     (17)

Where rand () is a random number generator function, Pos_x$(P_{i-1})$ is the x direction coordinate of the particle $P_{i-1}$, it may be possible that this exactly earlier particle of $P_i$ may not be best particle they we have to move $P_{i-2}$ and so on until we find best particle. Similarly pos_x$(P_{i+1})$ is X- direction position of particle $P_{i+1}$.

    E. Bounding Box formulation

To identify tracked target object usually a rectangle bounded box utilized. There are some pre defined algorithm exist for that purpose, but here we design our own bounded box based on QPSO particle position which will best suite our target tracking.

The main idea is whenever all particle in all swarm successfully converge for a particular image frame we find p number of particles which has smallest X – direction and smallest Y-Direction, those particle are close to (0,0) in our image space. This p value could be first 10 smallest particle, though it is entirely depends on application but as per our experiment goes it will be effective if we take first [10 – 20] smallest X,Y direction particles. Now take an average of those p- points, which will be our starting point for bounding box formation.

Let consider that particle is q,

$$q = \text{ceil} \left[ \{ p_1 + p_2 + p_3 \ldots\ldots p_{p-1} + p_p \} / p \right] \quad (19)$$

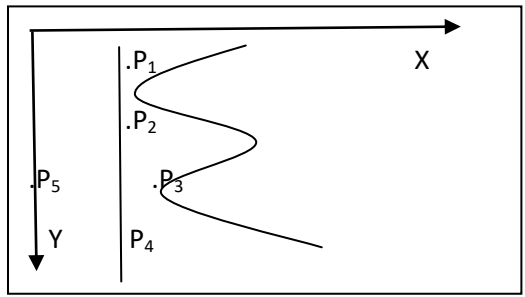

Figure - 3

This straight line shows we are taking average and point q is on the line. Now once we got the starting point (q) now we have to calculate length and breadth of the box. Length and breadth will be calculated as follows –

Length (L) = $\{ l_i \mid$ all $l_i$ are min(x- direction) and max (y- direction) $\}$ for i = 1,2,3…p

$$= \{ (l_1 + l_2 + l_3 + \ldots\ldots l_p) / p \} \quad (20)$$

Breadth (B) = $\{ b_i \mid$ all $b_i$ are min( y-direction) and max( x- direction) $\}$ for i=1,2,3…p

$$= \{ (b_1 + b_2 + b_3 + \ldots b_p) / p \} \quad (21)$$

So as now we got all 3 parameters we can draw the box as –

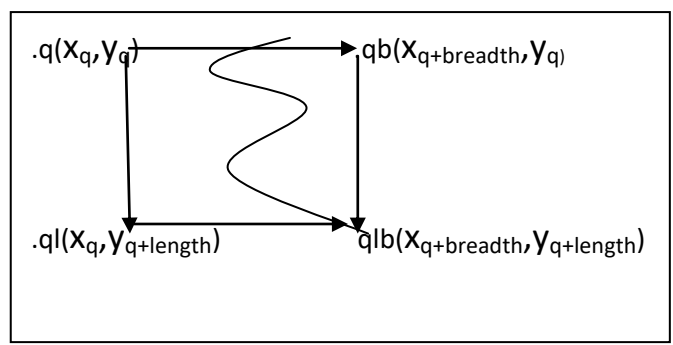

Figure – 4

As per figure (4) q is the starting point calculated using formula (14) and once we got L and B from formula (15) and (16) we can construct the box as –

$$\left.\begin{array}{l} q = (x_q, y_q) \\ ql = (x_q, y_{q+length}) \\ qb = (x_{q+breadth}, y_q) \\ qlb = (x_{q+breadth}, y_{q+length}) \end{array}\right\} \quad \text{four boundary position formula} \quad (22)$$

F. Algorithmic summary

If we summaries as a whole the QPSO object tracking algorithm that will be as following –

Procedure ObjectTrackingQPSO
    brpts ← calculate Breakpoints of target objects.
    dompts ← eliminate linear points and find dominant points using rule (6) and(7)
    nSwarm ← number of swarms
    ss ← define swarm size.
    for swarm ← 1 to nSwarm
        for $p_i$ to ss
            Initialize particles position.
            Initialize pBest and gBest.
            Compute Procedure FitnessComputeQPSO
        end for
    end for
    perform Bounding box calculation
    end Procedure ObjectTrackingQPSO

IV. Implementing QPSO in parallel environment :

One of the major feature in QPSO or PSO based approach is that if we can implement our algorithm in such a way that multiple swarms can run simultaneously then it will give a lot better results in terms of run time. Running parallel is inherent nature of PSO/ QPSO based algorithm. But implementing parallel environment is quite hard in sequential machine unless our code framework support that, fortunately MatLab do that.

MatLab has a special command for parallel run environment called "parfor", it is parallel implementation of "for" loop. By default "Parfor" will create 4 workers and distribute all its computation inside the loop into this 4 workers.so entire computation will be divided into 4 parts and each one will take 1/4rth parts of it, as a result its execution time is lot faster. But we have to remember here that whatever computation we have inside the loop must be independent, otherwise "Parfor" will not work.

In our work, we also make our code independent so that we can implement "Parfor" to create the parallel environment. Making entire code independent is rigours and unnecessary work, we only perform following work parallel –

1. Fitness computation
2. $P_i$ computation (Formula – 6)
3. Position computation (formula – 4,5)

In algorithmic structure it will look like this –

Procedure QPSO_Parallel ( nSwarm)
Begin
    Parfor 1 to nSwarm
        Procedure FitnessComputePSO();
        $P_i$ computation based on pBest and gBest;
        Position computation based on k values.;
    End Parfor
End

## V.    Experimental Result

The proposed algorithm for object tracking based on QPSO on different background is simulated by MatLab 2015a on a 64 bit PC with Intel i5 processor with 3 GHz speed. The image size of the frame 180 X 144. Static video is 20 sec duration whereas variable background is 13 sec duration.

All the experimental dataset has been taken form benchmark library created by *Yi Wu, Member, IEEE, Jongwoo Lim, Member, IEEE, and Ming-Hsuan Yang, Senior Member, IEEE*[1] work which will be available on http://pami.visual-tracking.net.

We first take frames from the video and store them in a separate location then use each frame in our algorithm for tracking.

### Static background

Our first experiment is on example 1 static background. In figure (5) a single person is coming towards the camera. Which multiple swarm successfully track that person and figure (6) shows same person with bounding box which is our own designed algorithm Consecutive figure shows how that person comes more and more close to the camera our algorithm successfully track that person.

Example 1:

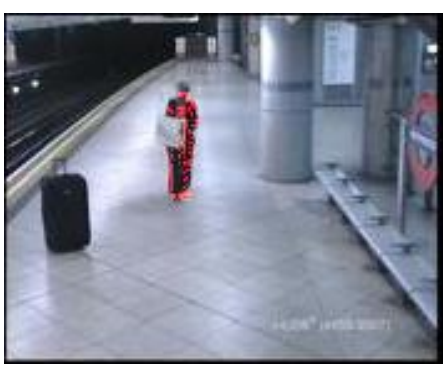 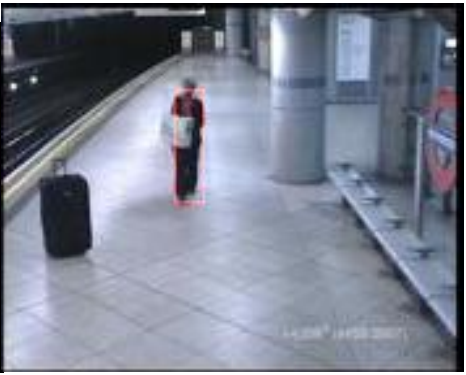

Figure – 5  Figure - 6

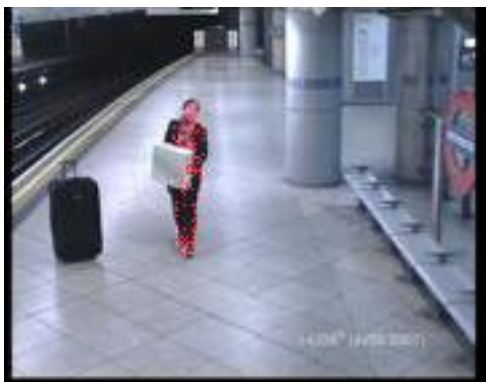 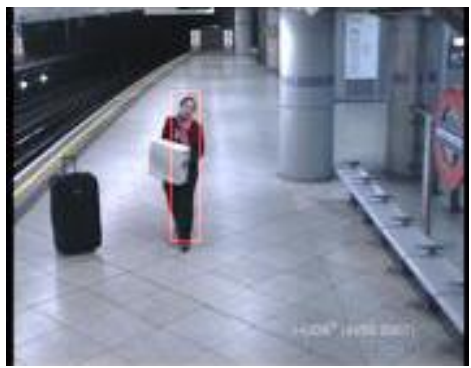

Figure – 7  Figure - 8

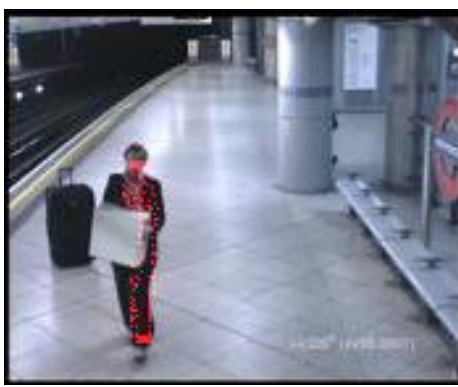 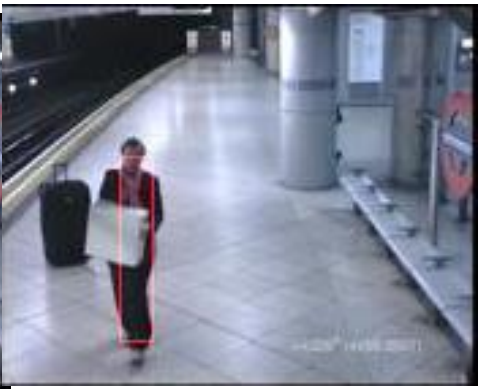

Figure – 9  Figure - 10

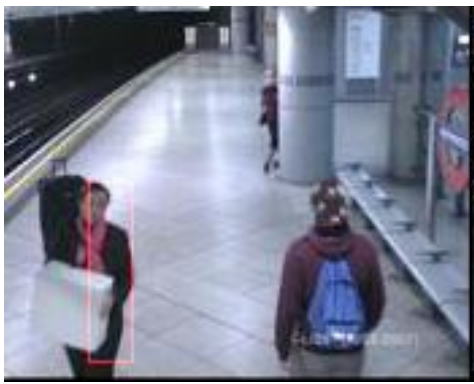 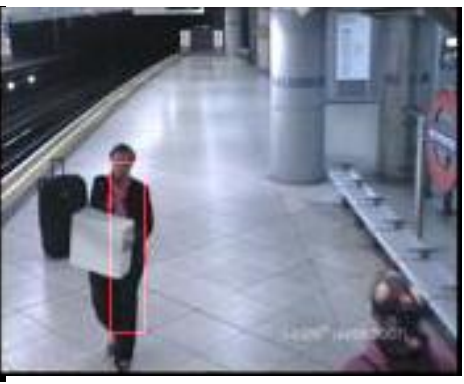

Figure – 11  Figure – 12

Figure (5) - (12) shows a sequence of frames of a single person moving towards a camera where background is static. All left side figure shows how swarms particles successfully track the figure as it moves .and all right side figure shows when we draw a bounding box using our QPSO based algorithm.

Example 2:

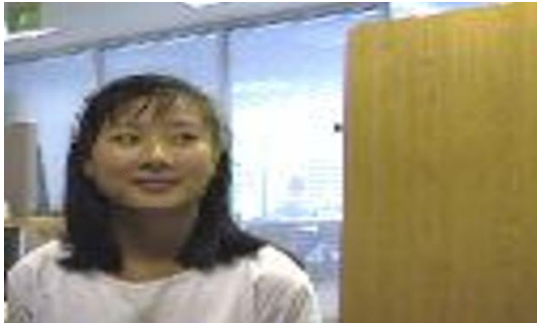  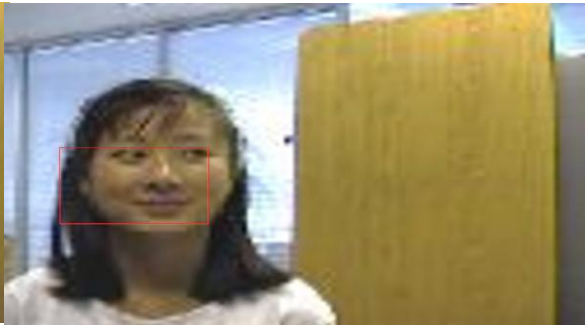

      Figure – 13　　　　　　　　　　　　　　　　Figure – 14

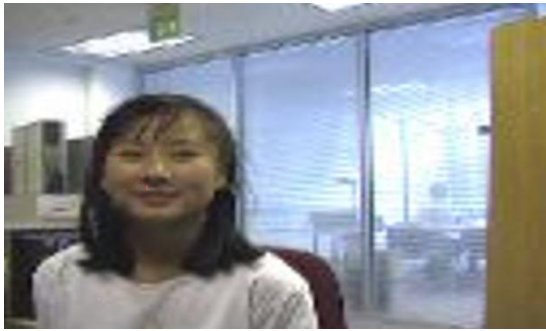  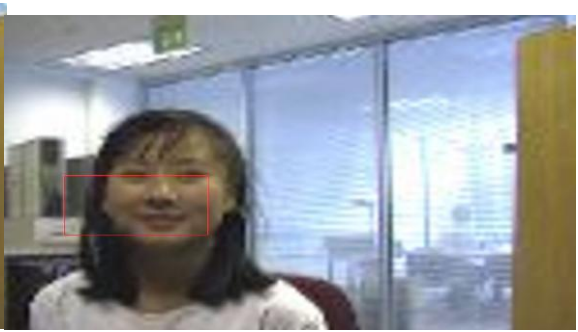

      Figure – 15　　　　　　　　　　　　　　　　Figure - 16

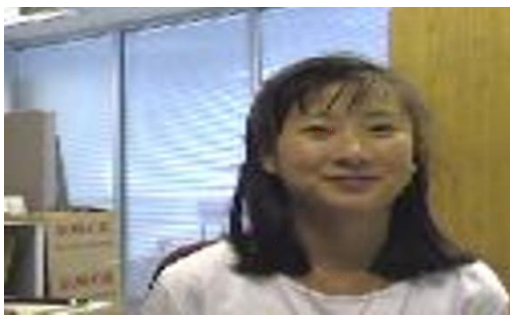  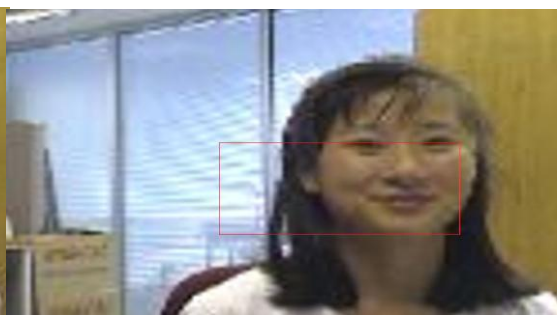

      Figure – 17　　　　　　　　　　　　　　　　Figure – 18

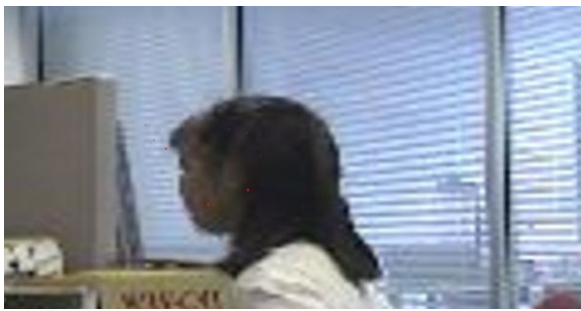  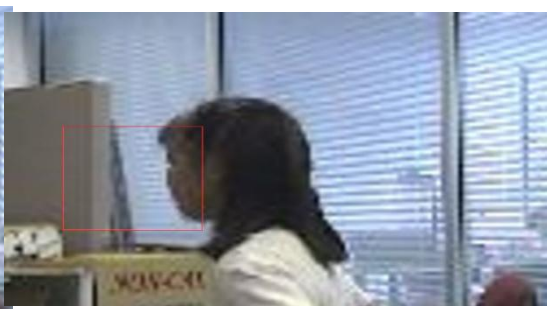

      Figure – 19　　　　　　　　　　　　　　　　Figure – 20

From figure- (13) – (20) A girl face is moving left and right, and back and forth , our algorithm successfully track the movement

Example -3

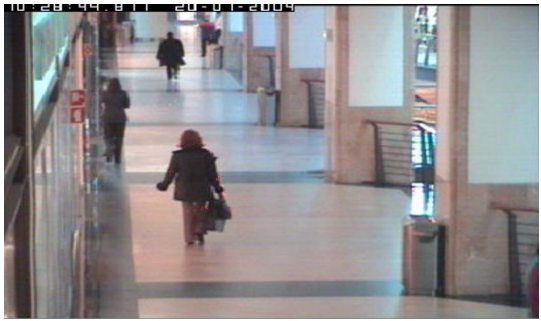

Figure -21

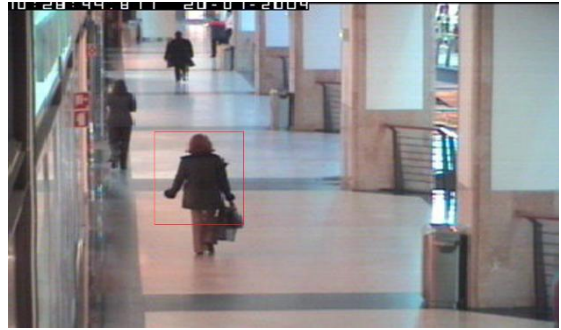

Figure- 22

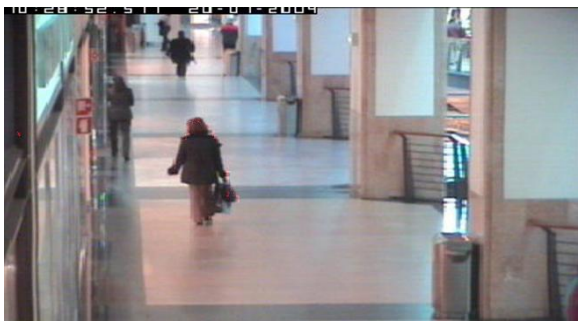

Figure -23

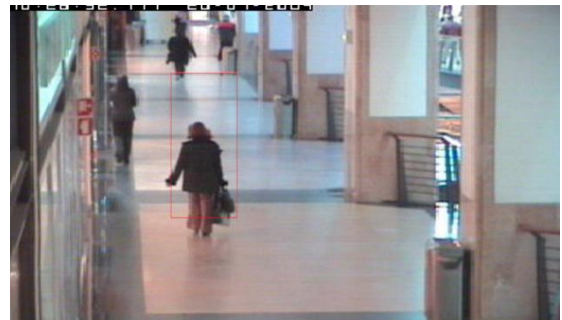

Figure -24

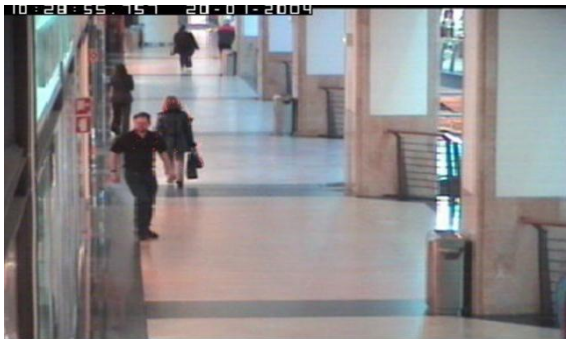

Figure -25

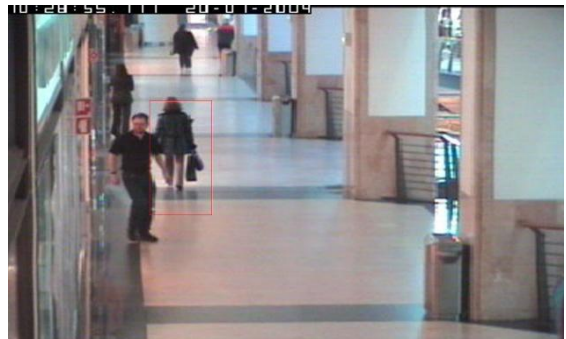

Figure -26

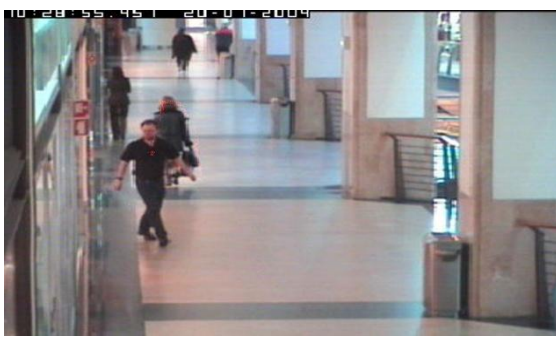

Figure -27

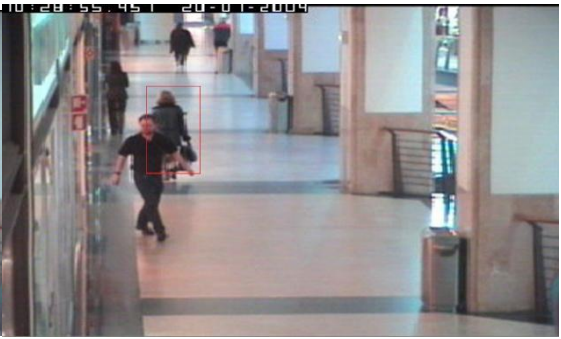

Figure -28

From figure- (21) – (28) A lady walking on a campus when she is obstructed by another man, or algorithm successfully track the lady during occlusion.

**Variable Background**

Now we experiment our algorithm on a video where background is moving with object, video is taken by a moving camera. Here we can see 3 persons moving from left to right with the camera, as two persons are very close to each other they are considered as a single tracking object, if one person moves far from another PSO will leave that person and focus on only that person who is under tracking, but if from beginning two separate persons are under tracking then if two person moves far from one another our algorithm will successfully track both the persons.

Example 1:

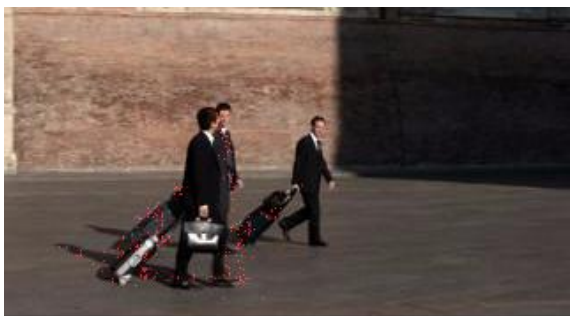

Figure – 29

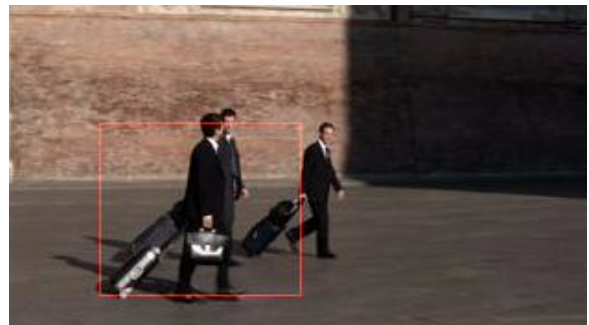

Figure - 30

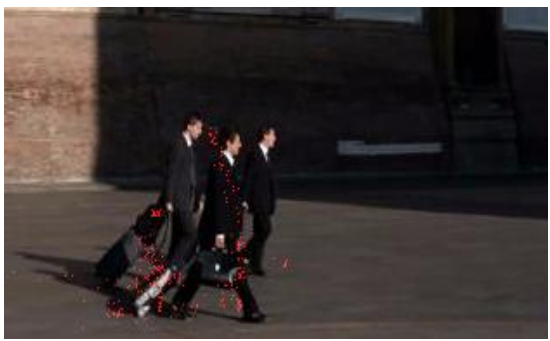

Figure – 31

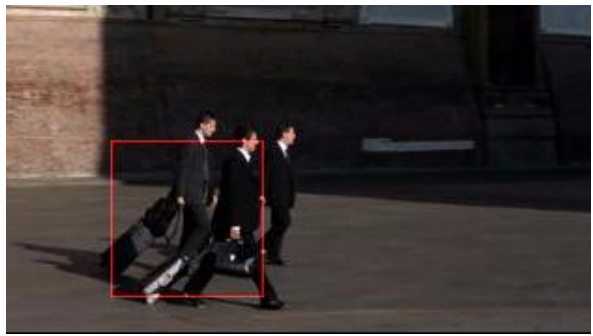

Figure – 32

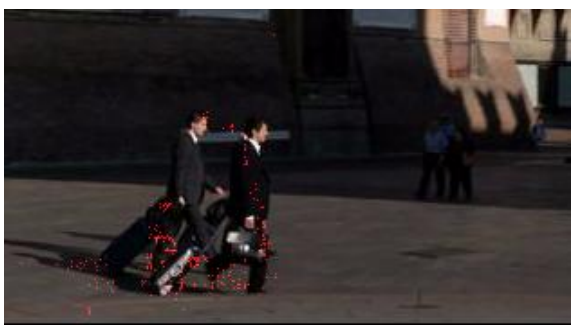

Figure -33

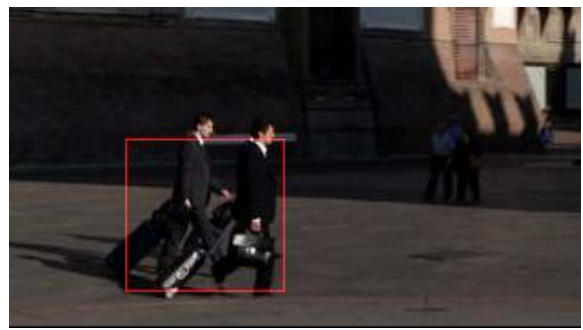

Figure -34

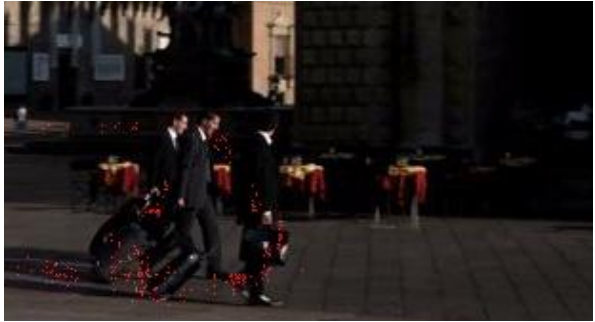 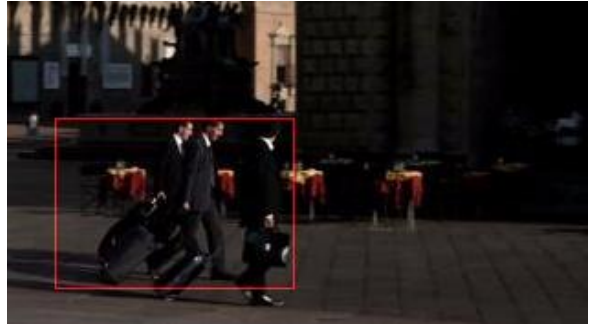

Figure – 35                                      Figure – 36

Form figure – (29) to (36) we are showing from left side swarms are successfully track the middle person on figure (29), and on right side we can see the same person is tracked on bounding box. Here we are tracking single person on not all three persons, they are comes inside the box due to their relative speed of movement. If we want to track them also it is possible by our algorithm.

Example 2:

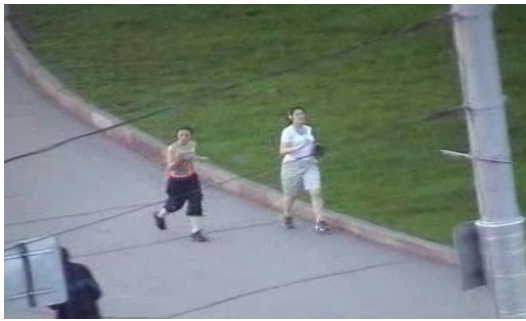 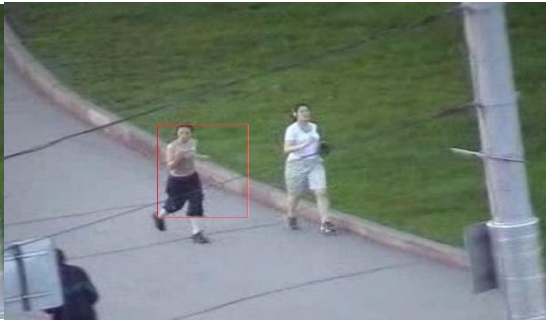

Figure – 37                                      Figure -38

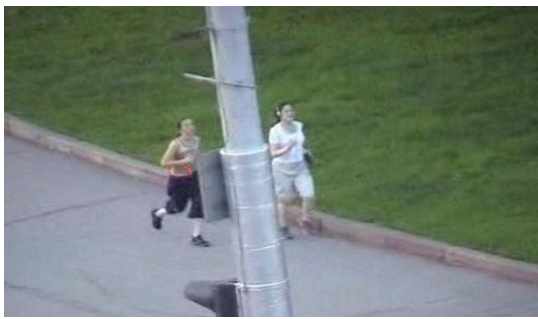 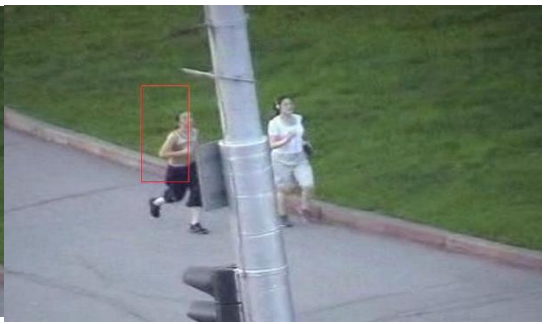

Figure -39                                       Figure -40

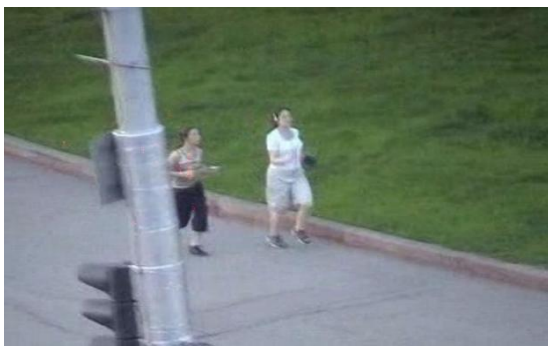 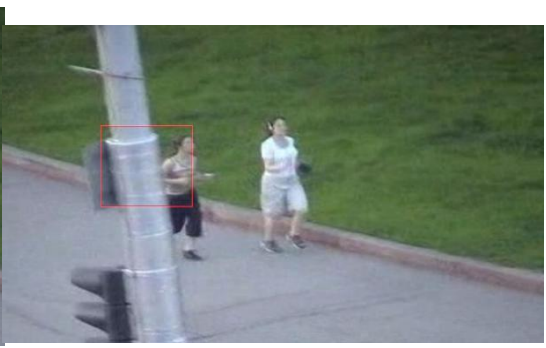

Figure -43 Figure 44

Form figure – (37) to (44) we are showing from left side swarms are successfully track the left jogger, where background is continuously changing and on right side we can see the same person is tracked on bounding box. Here traffic light post work as an occlusion, but out method able to track during occlusion.

Example – 3

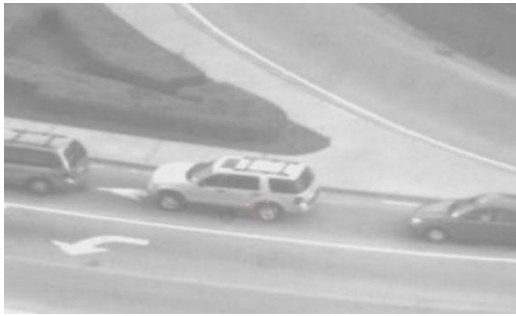
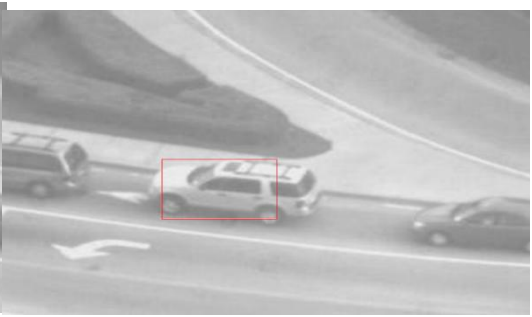

Figure -45     Figure -46

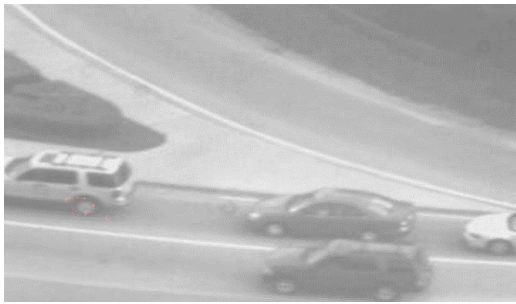
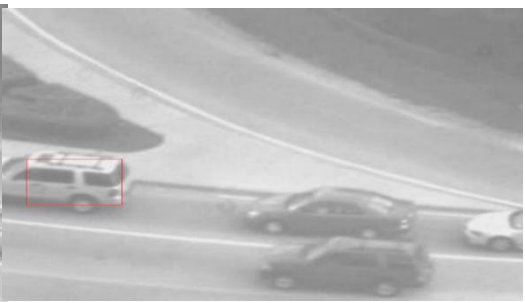

Figure - 47     Figure – 48

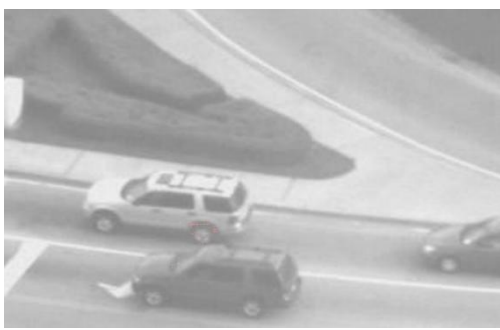
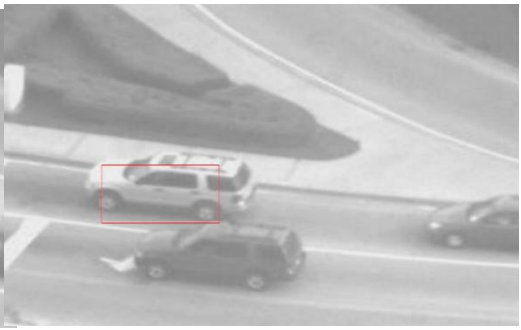

Figure -49     Figure -50

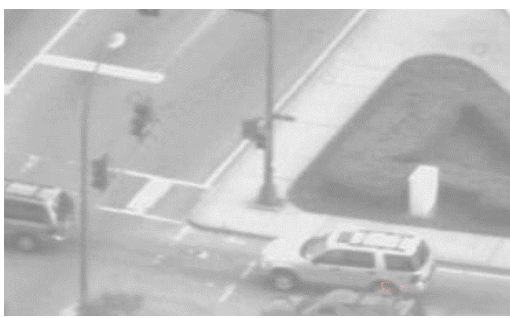
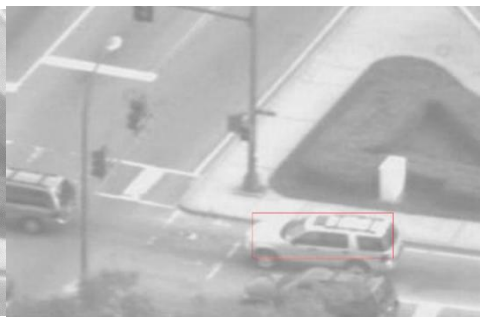

Figure – 51                                                                  Figure -52

Form figure – (45) to (52) this dynamic background video show a moving SUV car moving fast first then slow down when another car going to cross and den again slow down during traffic signal. The white SUV successfully tracked by our algorithm during all kind of movement.

## VI.   Overall Performance

A. **Performance comparison between basic PSO, Quantum PSO and Quantum PSO in parallel framework of our proposed algorithm**.­

Before going through all our test dataset and all respective results we specifically want to mention that we mainly focus on following parameters during test runs about performance analysis.

1. Iteration
2. Swarm size
3. Run time

| Test Environment | Test Data file | Test Parameters | Basic PSO particle run time | Basic PSO box run time | QPSO particle run time | QPSO box run time | QPSO_Para particle run time | QPSO_Para box run time |
|---|---|---|---|---|---|---|---|---|
| Static Background | Railway walking man.avi | Swarm Size | 25 | 25 | 15 | 15 | 7 | 7 |
| | | Iteration | 1000 | 1000 | 100 | 100 | 5 | 5 |
| | | Runtime | 1260 sec | 1221 sec | 1160 sec | **34 sec** | 1262 sec | 1149 sec |
| | Girl face moving.avi | Swarm Size | 25 | 25 | 15 | 15 | 7 | 7 |
| | | Iteration | 1000 | 1000 | 100 | 100 | 5 | 5 |
| | | Runtime | 4662 sec | 4431 sec | 3663 sec | **483 Sec** | 4561 sec | 4552 sec |
| | Lady waking on campus.avi | Swarm Size | 25 | 25 | 15 | 15 | 7 | 7 |
| | | Iteration | 1000 | 1000 | 85 | 85 | 7 | 7 |
| | | Runtime | 3389 sec | 3443 sec | 3300 sec | **289 sec** | 3210 sec | 3110 sec |

Table – 1: Comparison results obtained by PSO, Quantum PSO, and Quantum PSO in parallel for static Background in 3 different test file.

## Static Background Environment

The above table shows we perform test runs on a static background environment where we took 3 different .avi file. 4rth column shows basic PSO run time when only Swarm particles are marked on images like figure- 5 and next column shows same basic PSO run time when we formed a bounding box using our Bounding Box algorithm. Like figure -6. Next onwards columns are QPSO particle and box run time and QPSO particle and box run time when we use "Parfor" command for parallel framework. The above table shows when basic PSO took almost 21 minutes to run, Quantum PSO took 19 min for particle mapping but took 34 sec for bounding box design. Now if we consider Quantum PSO in parallel, its performance close to basic PSO but number of iteration and Swarm sizes are very less. The reason behind this the overhead for creating 4 workers is high then the computation inside the loop.so as a results its busy in doing overhead work rather than actual computation.

If we observe other two results QPSO gives so far the best performance results in bounding box designing which is our own designed algorithm. QPSO gives close to 96% performance improvement over basic PSO. Following graphs shows our performance improvement results.

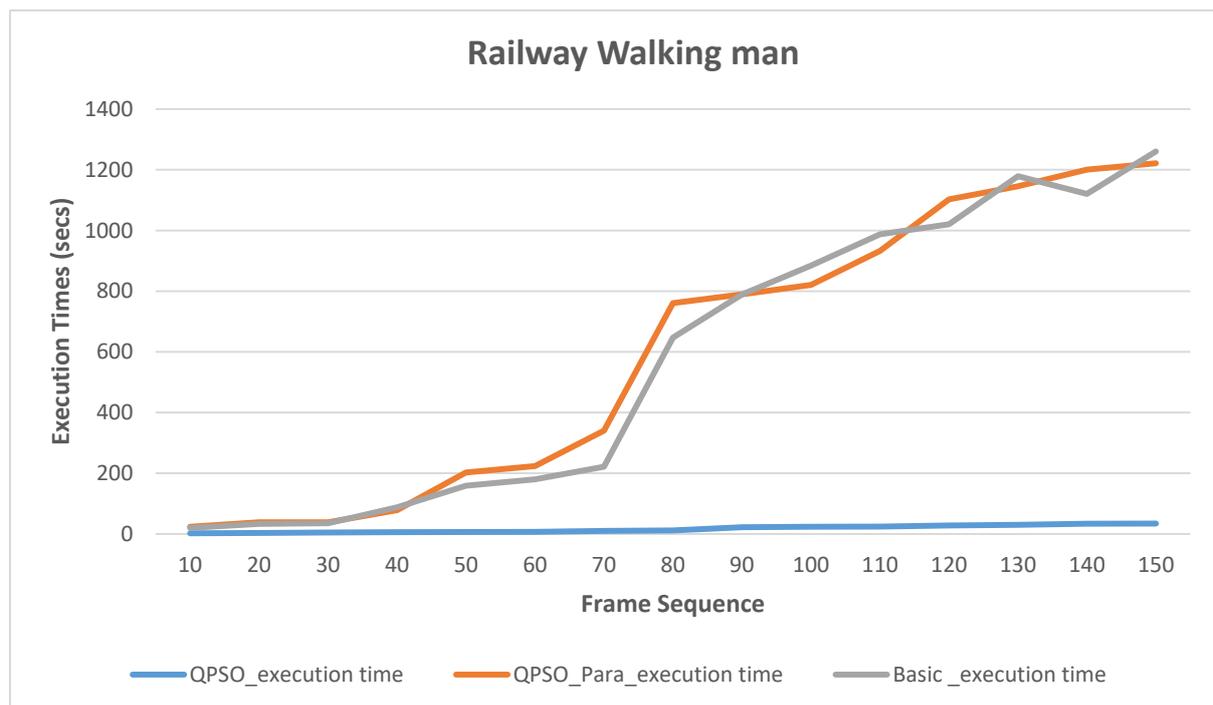

Figure – 53: performance comparison graph over basic PSO, Quantum PSO and Quantum PSO in parallel on Railway_Walking_man.avi video sequence.

In figure -53 graph in X- direction we have frame sequence and in Y- Direction we have Execution times. It is clearly observed that QPSO by far gives the best execution time result where other two increases their times proportionally with frame sequences. The above graph only shows one experiment graph(Railway_Walking_man.avi), others also gives quite same results, we intentionally keep aside creating same kind graph repeatedly, this above is more generalized performance graph on static background environment.

Table – 2: Comparison results obtained by PSO, Quantum PSO, and Quantum PSO in parallel for dynamic Background in 3 different test file.

### Dynamic Background Environment

| Test Environment | Test Data file | Test Parameters | Basic PSO particle run time | Basic PSO box run time | QPSO particle run time | QPSO box run time | QPSO_Para particle run time | QPSO_Para box run time |
|---|---|---|---|---|---|---|---|---|
| Dynamic Background | Three people walking.avi | Swarm Size | 35 | 35 | 20 | 20 | 10 | 10 |
| | | Iteration | 2000 | 2000 | 150 | 150 | 8 | 8 |
| | | Runtime | 8445 sec | 8223 sec | 7828 sec | **890 sec** | 8210 sec | 8093 sec |
| | Jogging.avi | Swarm Size | 35 | 35 | 20 | 20 | 10 | 10 |
| | | Iteration | 2000 | 2000 | 120 | 120 | 8 | 8 |
| | | Runtime | 7656 sec | 7234 sec | 6829 sec | **561 sec** | 6892 sec | 6726 sec |
| | white car moving on traffic signal.avi | Swarm Size | 35 | 35 | 20 | 20 | 10 | 10 |
| | | Iteration | 1600 | 1600 | 150 | 150 | 5 | 5 |
| | | Runtime | 8239 sec | 8120 sec | 5929 sec | **536 sec** | 7889 sec | 7656 sec |

Table -2 is in same format like table -1 but it is for those test files which works on dynamic environment.

We can observe that QPSO parallel took very less number of iteration and swarm size (10 iteration and 5-8 number of particles in a single swarm) again like static background, and QPSO gives better results again in terms of run times in comparison with others.

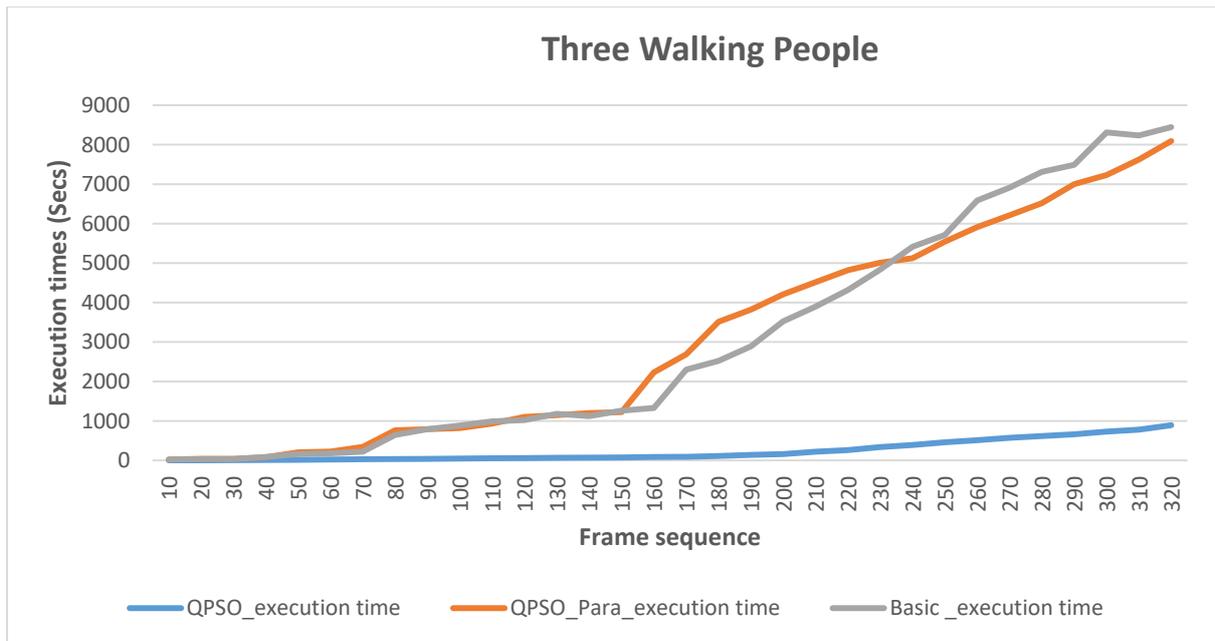

Figure – 53: performance comparison graph over basic PSO, Quantum PSO and Quantum PSO in parallel on three_walking_people.avi video sequence.

Same as earlier graph in figure-52, in dynamic framework number of frame sequence is higher than static one, as a result execution times takes longer time but it is evident from the above graph that QPSO gives us the far better result in object tracking in variable changing background.

B. Comparison in Multicore CPU Utilization for basic PSO, QPSO and QPSO parallel.

We also perform another comparison in terms of CPU utilization and Multicore utilization. The reason behind is that as we are using parallel framework our objective should be utilization of multicore and distribution of computation in all cores. We also want to observe how we can put more pressure on CPU and makes full utilization of CPY cycles.

We did this by opening task manager in our Windows – 7 64 bit operating system which installed in Intel Core i5 processor which works on 3.00GHz speed, it has 4 cores, all are utilized. Following figures shows how CPU are stressed and achieve their maximum performance and how all cores are properly used.

In following figure -54 we have seen CPU achieve 71% performance and first two cores are used core number 3 achieve much less performance during basic PSO run. In figure -55 during QPSO run CPU achieve 75% utilization with $3^{rd}$ core shows good improvement over basic PSO and in figure – 56 during QPSO in parallel run CPU achieve 88% performance with all cores are almost equally utilized, QPSO in parallel shows best CPU performance so far we tested.

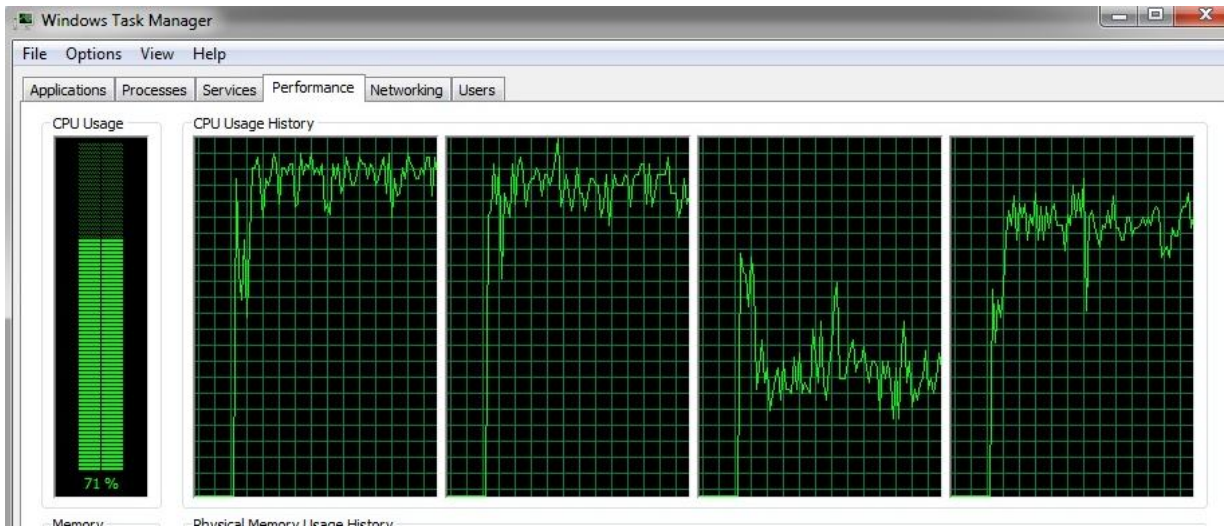

Figure – 54: CPU and Core Utilization during Basic PSO algorithm run

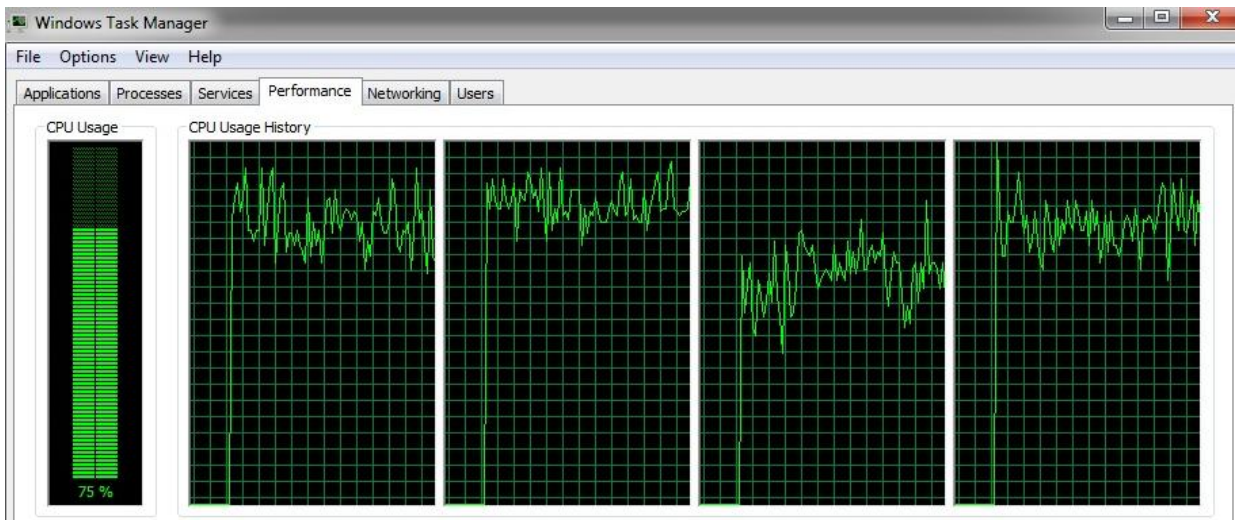

Figure – 55: CPU and Core Utilization during Quantum PSO algorithm run

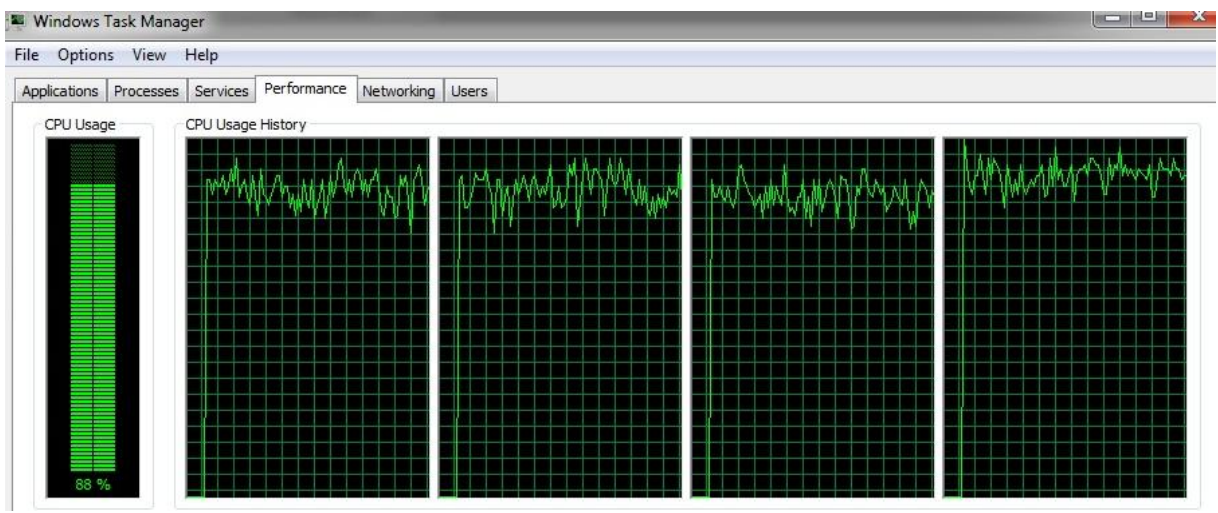

Figure – 56: CPU and Core Utilization during Quantum PSO in Parallel algorithm run

## VII. Conclusion and Future Work

Here in this paper we proposed a novel approach for object tracking by QPSO on both static and dynamic environment. The basic mechanism of this algorithm is introduced and proof of this approach is also provided by various experimental results. we also shows performance measurement on 3 different mechanism PSO, QPSO and QPSO-in-parallel. As we earlier discussed the uniqueness of this method is few things – I) Successful application of Dominant points in tracking purpose  II) a unified approach for  applying a singular algorithm in both the environment III) we achieve 95% faster method by applying QPSO approach over basic PSO. Our experimental results show how effectively our algorithm track objects throughout video frames and perform faster tracking.

We still can improve this method on various fine tuning grounds.  They are like 1) we can fine tune QPSO parameters like $\varphi$, $\beta$, $u$, $k$, mBest, more careful investigation will give us more accurate boundary tracking as well as bounding box designing. 2) curvature designing , we are initially assumes our curvature as a Euclidean Distance or considering a straight line, rather considering this if we fit B-Spline like curve which give us accuracy in curvature then our method will became more strong and provide more accurate boundary tracking. QPSO in parallel framework can be improved by including more CPU intensive computation inside "Parfor" loop which gives better result in tracking.

**Rajesh Misra** - is a lecturerin Information Technology at an Undergraduate college, India .prior teaching he worked in Software Industry as a Test Engineer in the field of Networking Protocol for 3 years. He received his Master in Technology degree From Calcutta University, India in 2010after receiving his B.Sc. and M.Sc. in Computer Science from same Institute in 2006 and 2008. His research interest includes Computer Vision, Particle Swarm Optimization, GeneticAlgorithm, Machine Learning, Image Processing.

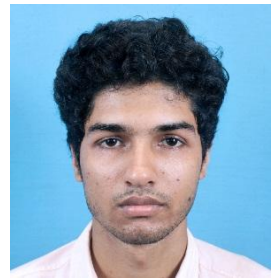

**Kumar S. Ray**- PhD, is a Professor in the Electronics and Communication Sciences Unit at Indian Statistical Institute, Kolkata, India. He is an alumnus of University of Bradford, UK. Prof. Ray was a member of task force committee of the Government of India, Department of Electronics (DoE/MIT), for the application of AI in power plants. He is the founder member of Indian Society for Fuzzy Mathematics and Information Processing (ISFUMIP) and member of Indian Unit for Pattern Recognition and Artificial Intelligence (IUPRAI).His current research interests include artificial intelligence, computer vision, common sense reasoning, soft computing, non-monotonic deductive database systems, and DNA computing. He is the author of two research monographs viz, Soft Computing Approach to Pattern Classification and Object Recognition, a unified concept, Springer, Network, and Polygonal Approximation and Scale-Space Analysis of closed digital curves, Apple Academic Press, Canada, 2013

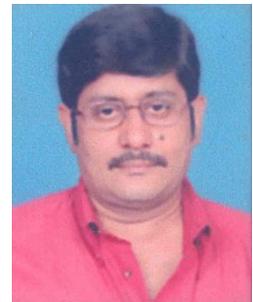